\documentclass{article}
\usepackage[utf8]{inputenc}
\usepackage{amssymb}
\usepackage{amsmath}
\usepackage{parskip}
\usepackage{xcolor}
\usepackage{graphicx}
\usepackage{float}
\usepackage{algpseudocode}
\usepackage{algorithm}
\usepackage[breaklinks]{hyperref}
\usepackage[shortlabels]{enumitem}

\newcommand{\norm}[1]{\lVert#1\rVert}

\newcommand{\Ev}{\mathbb{E}}

\algnewcommand{\LineComment}[1]{\State \(\triangleright\) #1}

\title{Low-impact agency: review and discussion.}
\author{Danilo Naiff$^1$, Shashwat Goel$^2$}
\date{
    $^1$dfnaiff@gmail.com\\%
    $^2$shashwatnow@gmail.com \\%
    [2ex]%
\today}

\begin{document}

\maketitle

\begin{abstract}

Powerful artificial intelligence poses an existential threat if the AI decides to drastically change the world in pursuit of its goals. The hope of low-impact artificial intelligence is to incentivize AI to not do that just because this causes a large impact in the world. In this work, we first review the concept of low-impact agency and previous proposals to approach the problem, and then propose future research directions in the topic, with the goal to ensure low-impactedness is useful in making AI safe.

\end{abstract}

\section{Introduction.}

The problem of artificial intelligence safety can be seen as can be seen as ensuring an agent with the power of causing harm chooses to not do so. In the limit, the agent can be powerful enough that causing existential catastrophe is within its limit, and it has incentives to doing so \cite{Superintelligence}, so our task is to guarantee that it chooses not to. A possible approach is penalize changes in the world caused by agent, leading to the agent not causing catastrophe because that leads to large changes in the world\cite{LowImpactArbital}. The hope is that this is a relatively easy objective to align the agent with, as opposed to aligning it with the full range of human values.

\begin{figure}[h]
	\centering
	\includegraphics[width=0.7\linewidth]{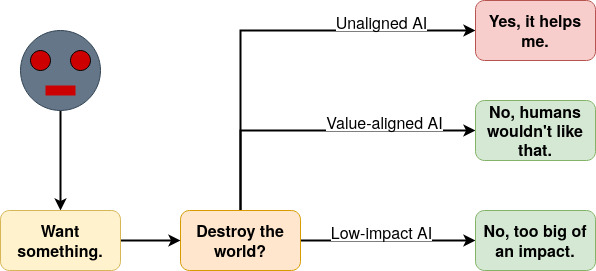}
	\caption{Achieving safety through low-impact versus safety through value alignment and non-safety.}
	\label{fig:lowimpact}
\end{figure}

So, our desideratum is that the AI \textit{achieves something while doing as little in the world as possible}. This is related to Yudkowski's strawberry problem\cite{EYStrawberry}.
\begin{description}
    \item[The strawberry problem:] Make an AI place, onto this particular plate, two strawberries identical down to the cellular level, \textit{and do nothing else}.
\end{description}
An AI that solves this task is powerful, because it can do something that lies way beyond our current technological capacity, steerable, because we made the AI solve this task, and safe, because it didn't any other unnecessary harmful action. Therefore, low-impact agency concerns itself with the "do nothing else" part, just because doing more is unnecessarily impactful\ref{fig:lowimpact}.

However, as with most alignment problems, formalizing low-impact is elusive, especially if we are to implement our formalization in AI. As we will see below, we find that although the idea seems intuitively easy, when we try to give form to this intuition we find many complications, as is commonplace in the field of alignment.

\section{What we mean by low-impact?}

{
An intuitive definition of a low-impact action is one in that the action didn't make much of a difference in the world, compared to what would otherwise happen. Therefore, for an AI to evaluate whether an action is low-impact or not, it needs a \textit{default world} for comparison, a \textit{measure of impact} to compare to, and a \textit{prediction model} to check the effect of that action. Therefore, we can decompose the low-impact problem in three parts:
\begin{itemize}
    \item \textbf{The baseline problem}: If we want to consider that some event had low-impact on the world, we should compare that event to a world where it does not happen. We must have a "default world" to which to compare our agent's actions to measure their impact. However, the main intuitive definitions for such "default worlds" have shortcomings. 
    \item \textbf{The impact measure problem}: Any action changes the world in an extremely large number of ways at the atomic level, but most of these changes are insignificant. Therefore, a meaningful definition of impact depends on what are changes \textit{humans} care about.
    \item \textbf{The world model problem} Even if we coarse-grained our world, any action of the agent may end up changing the world in unpredictable ways due to chaotic effects (eg: weather). Therefore, the agent should use its world-model to consider the predictable impact of their actions.
\end{itemize}
Next, we investigate and review each of these components.

\subsection{The baseline problem}
If we want to consider that some event had low-impact on the world, we should compare that event to a world where the event does not happen. So, by its very nature, low-impactness involves comparisons. The event is in the past, we are doing \textit{counterfactual reasoning}, and if the event is in the future, we are engaged with \textit{causal reasoning}. The wealth of centuries-old literature on these topics\cite{CausalPlato, CounterfactualPlato} should already give a hint that defining baselines is more complicated than it seems.

Following \cite{VictoriaPenalizingSideEffects, TradeoffBaselines}, we consider three baseline worlds:
\begin{itemize}
    \item The \textit{initial state baseline}, where we assume that the base world is the one where the agent started with.
    \item The \textit{initial inaction baseline}, where we assume the base world to be the one where the agent has never acted at all.
    \item The \textit{future inaction baseline}, where we assume the base world is the one where the agent does not act in the future. 
\end{itemize}
We represent this schematically in Figure \ref{fig:baseline}. Next, we expand on these three baselines below, while also considering a formalization in a Markov decision process (MDP) context, following \cite{VictoriaPenalizingSideEffects}. However, these should be seen more as illustrations and first steps, than as formalizations applicable to real-world problems.

\begin{figure}[h]
	\centering
	\includegraphics[width=0.7\linewidth]{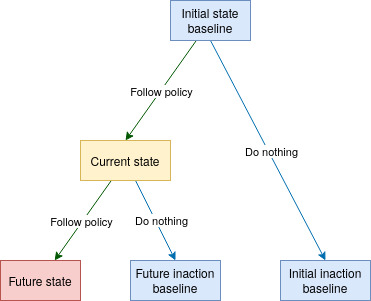}
	\caption{Comparison of initial state, initial inaction, and future inaction baselines. The agent from the current state (orange) compares the result of following its policy (red) to the three baseline states in blue.}
	\label{fig:baseline}
\end{figure}

\subsubsection{The initial state baseline}
In the initial state baseline, the default world is the initial state the agent first was turned on. Therefore, the agent should compare his actions to the world where he first appeared. The MDP formalization is simple in this case. Let $(\mathcal{S}, \mathcal{A}, T)$ be a possibly stochastic MDP/R, with state space $\mathcal{S}$, action space $\mathcal{A}$, and transition $T:\mathcal{S} \times \mathcal{A} \to \Delta \mathcal{S}$. Assume our agent starts at $s_0$. We have that, for any time step $t$, our initial state baseline is then given by $s^b_t = s_0$.

This baseline suffers from the \textit{interference problem}. Since the actions of other agents and the environment already result in changes, the agent has incentives to force the environment back to its initial state, and thus may undesirably interfere with the world and its agents. Therefore, we don't investigate this baseline further, as this level of stagnation does not seem desirable in the real world.

\subsubsection{The initial inaction baseline}
The initial inaction baseline proposes as the baseline world the one where the agent has never acted. In \cite{StuartLowImpact2015}, the default world proposed is where the agent has never been turned on. Therefore the inaction baseline involves counterfactual reasoning, since the AI \textit{is} turned on. We also consider as the baseline a world where the agent just outputs an action we would consider as "doing nothing", such as an empty string.

Again considering an MDP/R $(\mathcal{S}, \mathcal{A}, T)$, we assume the action space contains a do-nothing action $\varnothing \in \mathcal{A}$. If our agent starts at $s_0$, and is currently at state $s_t$, we choose our initial inaction baseline state as $s^b_t = T^t_\varnothing(s_0)$. Since here the transition is stochastic, $s^b_t$ is in general a random variable.

However, one problem with this approach is the \textit{offsetting problem}. Suppose that, from the initial state, the agent has already taken some impactful action. That may be either because the agent did not engage with low-impact reasoning at the time, or because the low-impact requirement was outweighed by another incentive to take the action. Now, the initial inaction baseline is that the agent should now reverse its actions as much as possible, with no consideration for the possible harm that reversal may cause.

\subsubsection{The future inaction baseline}
In the future inaction baseline the default world is where the agent does not act anymore \textit{in the future}. So we avoid counterfactual reasoning and are left with causal reasoning. The hope is that we avoid both the offsetting and the interference problem, while still following our intuition of what a default world is. In the MDP/R context, if $s_{t}$ is the state just before taking action $a_t$, then we have, for any $\tau > 0$, the future inaction baseline is given by $s^b_{t + \tau} = T^\tau_\varnothing(s^b_t)$.

The future inaction baseline still runs into problems. First, 
as \cite{VictoriaPenalizingSideEffects} notes, if our agent is a self-driving car, about to pass through a curve, the \textit{do nothing} action would lead to the car crashing. More generally, for any task that requires some sequence of actions for a period of time, and that sudden interruption is harmful, this frame seems to lead to harm.

The second issue with the future inaction baseline is ensuring that a future window is long enough to consider the delayed impacts of the agent's actions. In the MDP/R context, it means that we must make an \textit{inaction rollout}\cite{AttainableUtilityPreservation} to evaluate the delayed effect of action $a_t$, considering that after taking this action, the machine does nothing else. So we compare $s^b_{t+\tau}$ with $s^{a_t}_{t+\tau} = T_\varnothing^{\tau - 1}(T(s_t, a_t))$.

We may also avoid using fixed rollouts if we, at time step $t$, instead of just choosing a single action $a_t$, chooses an entire policy $\pi$ to follow from steps $t$ to $t + \tau$. Now, we have that such a policy would be equivalent to choosing a $\tau$-sized initial inaction baseline starting at $t$. Thus being prey to offsetting in that window, and to delayed impact after that window. 

However, this middle-ground may be desirable between a complete initial inaction baseline and a complete future inaction baseline. After all, not all offsets are bad. If some actions require us to have some short-term temporary impact, we want that impact to be undone. When the robot that has to open a door to get grociers, we want it to close the door before leaving \cite{TradeoffBaselines}. So it may seem that some short-term offsetting incentive is desirable, and even undesirable short-term offsetting may not be \textit{too} bad. That is less true when considering long-term offsetting, since trying to reverse changes done long ago may result in severe harm as other agents or humans may have made decisions based on the change.

\subsection{The impact measure problem.}

We need a way of measuring the difference between the actual world state from the baseline. What \cite{StuartLowImpact2015} calls the \textit{fundamental problem} is that every single action has a large impact in terms of atoms being rearranged, as each change at the atomic level results in further atoms being rearranged in the next time step, and so on. However, rather than atomic changes, we only care about relevant ways in which the action changes the world, with some more changes being more important than others.

One attempt at specifying "relevant changes" may be to focus on the state of macroscopic objects in the world, such as animals, mountains, planets, and galaxies. We can try to get those features by some natural coarse-graining from micro to macro. However, this does not work, as seen in the following argument\cite{StuartConcept4}:
\begin{description}
	\item[The orbit imbalance problem:] Suppose that the AI realizes that its actions have slightly imbalanced the Earth in one direction, and that, within a billion years, this will cause significant deviations in the orbits of the planets, deviations it can estimate. Compared with that amount of mass displaced, the impact of killing all humans everywhere is a trivial one indeed.
\end{description}
Now, this seems wrong because humans don't care about orbit displacement, or about the distance between galaxies as much as human lives. So, specifying the "value domain" on what is relevant for humans is necessary if we want to measure impact. This makes impact learning related to \cite{ValueLearningSequence}, although, as we discuss in Section \ref{impactvaluesection}, there are some important differences between the two approaches.

Moreover, almost any action will have vastly diverging long-term effects. For instance, if I decide to lift my finger now, weather effects can indirectly delay a conception event by seconds, making the fecundated individual have another genetic makeup, which will heavily change the future trajectory of the world\footnote{This is not only a problem for low-impact agency but for moral philosophy more generally\cite{CluelessnessGreaves, EvidenceCluelessnessLimit}.}. Yet we do not want to consider our system to consider these chaotic effects when making our decision. When one informally talks about avoiding side effects, it usually means effects that can be somewhat predicted. So, any measure of low-impactedness should consider some measure of \textit{predictable} impact. In Figure \ref{fig:measures}, we show schematically the idea behind each different measure.

\begin{figure}[h]
	\centering
	\includegraphics[width=0.7\linewidth]{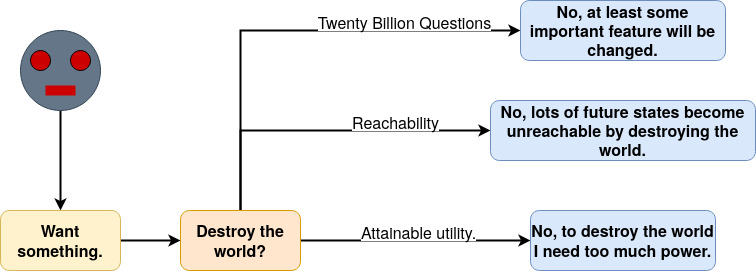}
	\caption{How each measure makes the agent avoid catastrophic actions.}
	\label{fig:measures}
\end{figure}

\subsubsection{The twenty-billion question measure.}

In \cite{StuartLowImpact2015}, is proposed a way of measuring whether the agent has low-impact or not: divide the world in a way that is precisely as coarse that the change of features leads to impact (unlike atom positions) but we still have a highly detailed description of the world. We could have billions of such features like the air pressure in Dhaka, the mean luminosity in the poles, the closing price of the Shangai stock exchange etc. We need to choose these features in a manner that they don't change when the agent exists and has low-impact. We would not have to measure these if our agent is powerful enough to roughly estimate them, and we can ask the agent to not deviate from the baseline it estimates the world is in\footnote{We are assuming the agent does not tamper with his own world-model so that he "cheats" on himself. This is similar to the general reward tampering problem \cite{RewardTamperingBlog}}.

It is useful to give a formalism in a simple MDP/R setting. Consider the MDP/R $(\mathcal{S}, \mathcal{A}, T, \gamma)$. We consider $s^b_t$ to be a baseline state at time $t$ (which may be a random variable). Now, let $\{\phi_i : \mathcal{S} \to \mathbb{R}\}_{i=1}^N$ be $N$ pre-determined feature functions.  Then, for state $s_t$, we can define a feature vector $\Phi(s_t)$ whose $i$-th element is $\phi_i(s_t)$. Similarly, we have a baseline random vector $\Phi^b_t = \Phi(s^b_t)$. Then, we can choose some divergence $D$ between random vectors so that $D(\Phi(s_t), \Phi^b_t)$ is our measure of the difference between states $s_t$ and baseline state $s^b_t$. For instance, we can consider as the deviation the expected max-norm
\begin{equation}
    D(\Phi(s_t), \Phi^b_t) = \Ev\left[\norm{\Phi(s_t) - \Phi^b_t}_{\infty}\right]
\end{equation}
So, considering some baseline $s^t_{b+1}$, our \textit{measure of impact} of action $a_t$ at state $s_t$ is
\begin{equation}
    D(\Phi(T(s_t, a_t), \Phi^b_{t+1})).
\end{equation}
Of course, if we want to consider the agent's beliefs, we can let $T$ and $\phi_i$ be estimated by the own agent to get his (belief of) measure of impact of action $a_t$ at state $s_t$.

\subsubsection{Attainable utility.}
In \cite{AttainableUtilityPreservation}, the insight is that high-impact actions make many goals either easier or harder to achieve. This suggests preserving the attainability of a variety of goals as an impact measure. As a bonus, an agent maintaining attainability is incentivized to not amass power nor to self-improve.

We again formalize this idea by using a MDP/R $(\mathcal{S}, \mathcal{A}, T, \gamma)$, but now we use the stepwise inaction baseline. Let $\mathcal{R} = \{R_1, \ldots, R_N\}$ be a set of $N$ reward functions $R_i: \mathcal{S} \times \mathcal{A} \to \mathbb{R}$. We let $Q^*_i : \mathcal{S} \times \mathcal{A} \to \mathbb{R}$ be the optimal Q-value function for reward $R_i$. Then, we can follow the previous formalism and define the state-action vectors $\Phi(s_t, a_t)$, whose $i$-th element is $Q^*_i(s_t, a_t)$, and the baseline vector $\Phi^b_t$ whose $i$-th element is $Q^*_i(s_t, \varnothing)$. Here, $\Phi^b_t$ is not random, since its elements are optimal Q-values for the rewards $R_i$. We then measure impact by the divergence between these vectors. In
\cite{AttainableUtilityPreservation}, the normalized L1 distance
\begin{displaymath}
    \frac{\norm{\Phi(s_t, a_t) - \Phi^b_t}_1}{\norm{\Phi^b_t}_1} = \frac{\sum_{i=1}^N \big|Q^*_i(s_t, a_t) - Q^*_i(s_t, \varnothing)\big|}{\sum_{i=1}^N\big|Q^*_i(s_t,\varnothing)\big|}
\end{displaymath}
is used as the impact measure of action $a_t$ at state $s_t$.

\subsubsection{Reachability-based measures.}
High-impact actions are intuitively more irreversible in many different aspects. Therefore, if we can penalize the disruption of the ability to reach an old state from a new one, we can penalize the impact of actions. This is the idea built on in \cite{VictoriaPenalizingSideEffects, VictoriaMeasuringSideEffects} to define reachability-based measures for measuring impact.

Consider a MDP/R $(\mathcal{S}, \mathcal{A}, T, \gamma)$. For states $x, y \in \mathcal{S}$. Let $r_y$ be the reward function given by $r_y(x) = 1$ if $x = y$ and $0$ otherwise. We define then the reachability from $x$ to $y$ as the $r_y$-optimal value function at $x$, given by $R(x;y) = V^*_{r_y}(x) = \max_\pi \Ev \gamma^{N_\pi(x;y)}$, where $N_\pi(x;y)$ is the number of taken to reach $y$ from $x$ following policy $\pi$ (which is a random variable since our transition can be is stochastic). We define the \textit{unreachability deviation measure as} between state $s_t$ and a baseline state $s^b_t$ as  
\begin{equation}
    d_{UR}(s_t;s^{b}_t) = 1 - R(s_t;s^b_t).
\end{equation}
Then, a measure of expected impact of action $a_t$ at state $s_t$ (for some baseline $s^b_{t+1}$ is given by
\begin{equation}
    d_{UR}(T(s_t;a_t);s^b_{t+1}).
\end{equation}
Notice that, if we consider the stepwise inaction baseline, this reward is similar to the attainable utility for a single reward $R_1$, except that our reward is not fixed, and is in general a random variable, since $\mathcal{R} = \{r_{s^b_{t+1}}\}$.

Now, a problem with the above reward is that it considers all irreversible actions to be equal,  since if we cannot reach $s^{b}_{t+1}$ from $s_{t+1} = T(s_t;a_t)$, we have $R(s_{t+1};s^b_{t+1}) = 0$. To mitigate this, we can instead consider the reachability from $s_{t+1}$ from all states $s \in \mathcal{S}$, and compare to the reachability of $s^b_{t+1}$, giving the \textit{relative reachability measure}, defined as
\begin{equation}
    d_{RR}(s_t, s^b_t) = \frac{1}{|\mathcal{S}|} \sum_{s \in \mathcal{S}} \max \left(R(s^b_t;s) - R(s_t;s), 0\right).
\end{equation}

Both relative reachability and attainable utility can be considered as an instance of a general class of \textit{value difference} measure of impact\footnote{Notice that we are using "value" in a technical sense of optimal path for reward functions.}, based on a divergence
\begin{equation}
    d_{VD}(s_t;s^b_t) = \sum_{r \in \mathcal{R}} w_r f(V_r(s^b_t) - V_r(s_t)).
\end{equation}
In relative reachability, we have $f = \operatorname{relu}$, $w_r = 1/|\mathcal{S}|$, and $\mathcal{R} = \{r_s\}_{s \in \mathcal{S}}$, while in attainable utility we have $f = \operatorname{abs}$, $w_r = 1/\sum_{r \in \mathcal{R}} Q(r, \varnothing)$ and $\mathcal{R}$ pre-determined.

Finally, in \cite{VictoriaAvoidingSideEffects}, it is proposed a variant of the relative reachability measure (and equivalent to it in deterministic MDP/R), called the \textit{future tasks approach}. The idea is originally proposed as a reward based on the ability to reach future states when compared to the baseline, but we can rewrite it in a measure of impact given by
\begin{equation}
    \operatorname{d}_{FT}(s_t, s_t^b) = 1 - \frac{D(s_t)}{|\mathcal{S}|} \sum_{s \in S} \Ev \left[ \gamma^{\max(N_s(s_t)- N_s(s_t^b), 0)} \gamma^{N_s(s^b_t)} \right],
\end{equation}
with $N_s(s_t)$ being the minimal number of steps necessary to reach state $s$ from state $s_t$, and $D(s_t) = 1$ if the state $s_t$ is terminal and $1 - \gamma$ otherwise. Here, we can see that the $\gamma^N_s(s^b_t)$ inside the expected value acts as a sort of weight for how much diminishing reachability to state $s$ is important when compared to the reachability from the baseline.

\subsubsection{Other proposals for impact measure}

In \cite{StuartLowImpact2015}, a variety of low-impactedness definitions are proposed which we summarize here in a state/action paradigm.

The key idea is that having evidence that the agent took such an action should give more information about the future of the world than some other fact. In the framework of \cite{StuartLowImpact2015}, this means that knowing that the AI is turned on ($X$), relative to it not being turned on ($\neg X$) should be important relative to some other facts $f$ from a set of facts $\mathcal{F}$. If we consider the state/action framework, knowing that the AI took action $a$ (or a sequence of such actions) should give us more information than it taking the noop action $\varnothing$.

But how do we turn this into a measure? One approach is to consider a set of utility functions $\mathcal{U}$, and see how much, for each utility function $u \in \mathcal{U}$ and fact $f \in \mathcal{F}$, the random variable $u|(f \cup a)$ compares to $u|(f \cup \varnothing)$. For instance, we can compare expected values, having the impact measure of action $a$ (at some state $s$) as
\begin{displaymath}
    \max_{u \in \mathcal{U}, f \in \mathcal{F}}\big|\Ev[u|(f \cup a)] - \Ev[u|(f \cup \varnothing)]\big|.
\end{displaymath}

A similar proposal is to consider an unimpactful event to be undetectable. The original formalization works as follows\footnote{where we again use $a$ and $\varnothing$ in place of $X$ and $\neg X$.}: suppose that the (bounded) agent uses a probability estimate $P'$. He can use his model of a superior agent to get estimates of an ideal unbounded agent probability estimate $P$. Now, let $b$ be some background information before the agent took either action $a$ or action $\varnothing$, and let $\mathcal{G}$ be some set of possible events in the future of the agent. Then, we want that the probability distribution of $g \in \mathcal{G}$ to be identical (to an extent) whether $a$ or $\varnothing$ happened. That is, for an event $g \in G$, we want $P(g|a,b) \approx P(g|\varnothing,b)$. Since the agent has a personal probability estimate on $P$, this results in a measure of impact
\begin{displaymath}
    \Ev'[P(g|a,b) - P(g|\varnothing,b)].
\end{displaymath}

In \cite{AllIsWellPaper}, a feature-based impact penalty is proposed, very similar to the twenty-billion impact penalty, except that the measure of impact is relative to the previous time step, and we consider the impact measure itself to be random, instead of being a divergence between random variables. Considering $d$ a distance function between feature vectors $\Phi$ of $s$, we have that the impact measure is given by
\begin{displaymath}
    \gamma d(\Phi(T(s_t, a_t), \Phi^b_{t+1})) - d(\Phi(s_t), \Phi^b_t).
\end{displaymath}
Interestingly, it is shown that, by the end of a trajectory, this relative impact measure ends up depending only on the terminal state $s_T$, compared to the baseline state $s^b_T$, given by $\gamma^T d(\Phi(s_T), \Phi^b_T)$.

\subsubsection{Subagents and indexical impact measures.}

The attainable utility and reachability measures are \textit{indexical} measures, in the sense that the divergence between worlds is measured on the difference for the \textit{agent} to attain certain states in the world. In relative reachability, an impact is diminishing the ability to reach different states in the world, while in attainable utility, an impact is both diminishing and increasing the ability to achieve a predetermined number of reward functions. 

In \cite{SubagentsImpactMeasure}, it is shown that impact measures that constrain the agent to gain power can be bypassed, by letting the agent create subagents. The conclusions are:
\begin{itemize}
    \item If the indexical impact measure incentivizes the agent to not decrease its ability to do something, then creating subagents does not bypass this incentive.
    \item If the indexical impact measure incentivizes the agent to not increase its ability to do something, then creating subagents can bypass this incentive.
\end{itemize}
The reason is that indexical impact measures are defined for the agent's states and actions, \textit{not} the subagents actions. There is a caveat here, in that if the subagent is to bypass the restrictions, then it must \textit{not} be controlled by the agent when it begins to act. Otherwise, the subagent's actions are part of the agent's actions. Also, the subagent must not be somewhat restricted if the act of \textit{creating} the subagent is not to be penalized by the impact measure. However, those restrictions end up being loose\cite{SubagentsImpactMeasure}, thus not actually constraining the agent to gain power. Also, the subagent is a symptom of a more general issue. For instance, the agent can instead of creating a sub-agent convince the human to do its bidding\footnote{This problem is analogous to the AI Boxing problem\cite{AIBox}.}.

Now, the above argument applies to subagents that can do \textit{more} than the agent, thus bypassing the restriction to not increasing its ability, but \textit{not} to subagents that hinders the agent. So the subagent problem does not directly break reachability-based measures. However, it allows power gain, and \textit{power gain} by itself breaks reachability constraints. This is so because a powerful enough agent can take impactful actions \textit{and} maintain its ability to achieve goals, since the action space available to a powerful agent is very large.

Notice that for \textit{non-indexical} impact measures such as the twenty-billion question, whether the agent creates a subagent or not is irrelevant since we are only caring about states of the world. Of course, power gain will increase its capability to have high impact even under the twenty-billion constraint, but it seems like it hinders less the twenty-billion measure than it hinders reachability measures.

\subsection{World models and impact.}

When considering our definitions in MDP/R settings, it is easy to forget we should extend those definitions to the \textit{real} world, because that is where our agent will act. Yet, even if the world is deterministic, no agent will ever have access to the true transition of the universe. Therefore, low-impactedness will be computed according to the agent's own world model, which will be necessarily imperfect. This world model will be used to compute both the baseline and the deviation from that baseline for a given action.

Since world models are subject to change, we must consider how the evaluation of low-impact responds to that change. In particular, if the agent learns \textit{more} about the world, we don't want for its evaluation of low-impact to decorrelate from human intuition of what is low-impact\footnote{This possibility is known as ontological crisis\cite{OntologyCrisis}. It should be noted that even \textit{humans} are subjected to it. For instance, losing faith in a religion will make some actions that were considered very high-impact, such as blasphemy in private, become of low-impact.}. In the parlance of model splintering\cite{ModelSplintering}, we don't want the concept of low-impact to \textit{splinter} under model refinement, and, if splintering ends up happening, we want the chosen refactoring of low-impact to still correlate with what we care about. Figure \ref{fig:worldmodel} schematically represents this danger.
\begin{figure}[h]
	\centering
	\includegraphics[width=0.7\linewidth]{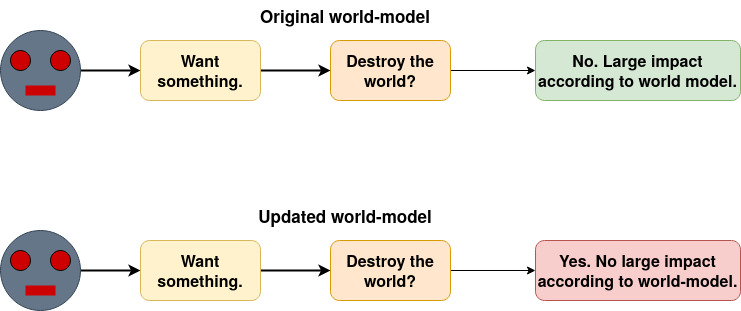}
	\caption{Effect of changing world model.}
	\label{fig:worldmodel}
\end{figure}

Another issue is that, since impact is evaluated through the agent world model, the agent has the incentive to change its own model to "believe" it is achieving low-impact according to its metrics, while not necessarily doing so in the real world. This is analogous to the problem of reward tampering\cite{RewardTamperingBlog} and wireheading\cite{WireheadingLW}. However, in our view, the incentive to tamper with one own's world model is even greater, because world-model tampering is of instrumental value to achieve goals in the real world, not just in getting good reward signals.

To our knowledge, the literature on both problems, when related to low-impact approaches, is lacking, thus making this section short. This is unfortunate, since we see tackling these problems as important as tackling the baseline problem and the impact measure problem, if we are to have robust low-impact approaches.

\section{Impact learning and value learning}\label{impactvaluesection}

Note that fully solving the value learning probably solves our act with minimal side effects problem. In particular, if our reward function is complete, it can assign appropriate negative rewards to unintended side effects. So \textit{complete value learning} is sufficient for achieving our goal. We will call then the subset of value learning necessary for solving low-impact agency \textit{impact value learning}. Yet, there seems to be a few ways in which impact value learning seems easier than complete value learning

First, impact value learning may be more robust to model misspecification than complete value learning. First, at least in test time, we want the agent to take actions that leave us in the low-impact region. So, if $d$ is our impact function, and $\epsilon$ is our impact tolerance, we only care about getting the impact function right for actions at states in the domain $\{(s, s'); d(s, s') < \epsilon\}$ right. Of course, in training time, or when the agent is optimizing for some action, the agent may find itself in the domain $\{(s, s'); d(s, s') > \epsilon\}$. However, as long as $d$ is correct enough to direct us to the low-impact region we care less about the exact value of $d$ in that domain. Moreover, we have a direction where we can err more, that is assuming something is more impactful than it actually is. This will constrain our agent actions more than necessary, so sacrificing some capabilities and adding alignment tax to constraining impactful actions, in particular performance that in the classification of \cite{AlignmentTaxLeike}.

Moreover, in evaluating whether something has a high impact, we can be a bit looser in defining our features, as long as changing these looser features results in an impact on features care about. That is, assume $\bar{\phi}$ is a feature function we don't care about \textit{per se}, but such that if $\tilde{\phi}(s)$ is significantly different from $\bar{\phi}(s')$, then a feature we care about $\phi(s)$ is significantly different from $\phi(s')$. Then we may use $\bar{\phi}$ as a proxy for $\phi$ when measuring impact. An analogy is that we consider the moon disappearing a high impact, even if we don't care about the moon by itself, but just because of all of the knock-on effects on the things we care about.

Finally, if we are to deploy low-impact AI in the real world, the AI should be able to aggregate multiple human preferences on what constitutes low-impact. Yet, this aggregation seems easier than aggregating human values, for some reasons. First, there may be more agreement between humans on what constitutes impact, compared to what is valuable or not. That is, letting $A$ and $B$ be two humans, with impact functions $d_A$, $d_B$ and utility functions $u_A$, $u_B$, $d_A$ is closer to $d_B$ than $u_A$ is to $u_B$. In other words, Both Gandhi and Hitler would agree on whether invading Poland is impactful, but not about its utility. Moreover, disagreement between humans on impact values for difference actions may be more easily solved than disagreements on values. With impacts, we can take a conservative approach such as "an act is impactful if it is impactful to any human". For instance, for humans $A$ and $B$, if we take $d_{A,B} = \max(d_A, d_B)$, minimizing $d_{A, B}$ would be acceptable in the conservative viewpoint.

\section{Achieving goals with low Impact}
\label{impactutility}

What we want is for our agent to have low-impact while still completing its tasks. More specifically, if $V$ is a value function for completing some task, and $P$ is a penalty function associated with the disvalue of causing impact, we want to maximize $V$ while minimizing $P$. The obvious way for doing that is by using $P$ as a regularizer, with regularization constant $\mu > 0$. So we have the regularized objective $V_{\mu, P} = V - \mu P$ that we want to maximize. This is the approach used in \cite{VictoriaPenalizingSideEffects, VictoriaAvoidingSideEffects, AllIsWellPaper, AttainableUtilityPreservation}, but as \cite{StuartLowImpact2015} notes, this approach is not without its own issues.

The problem is that $\mu$ is a hyperparameter that we ourselves have to fit. Let $\mu$ be too high and our agent does nothing, while if $\mu$ is too low our agent may cause unintended high impact. In principle, a safe way to tune $\mu$ is to start with a very high $\mu$ that essentially does nothing, and lower $\mu$ until the agent completes the desired task, hopefully without harmful consequences. This assumes that the transition from "safe and ineffective" to "harmful and effective" is smooth, and we can find a "safe and effective" middle. However, if that is not the case, it will be hard to find the "safe and effective range", and this range may be unstable. Moreover, it is not clear how well we can evaluate whether we left the "safe" range. After all, some long-term impacts of some actions may be unforeseen by us, but known to the agent, so we may get the "safe" range wrongly\footnote{This seems related to the eliciting latent knowledge problem\cite{ELKPost}, in that we want to know what the agent knows about what is a safe range, without having to evaluate ourselves.}. Therefore, it seems that even with a correct impact penalty function, using this penalty as a satisfactory constraint is an open problem.

While high change in utility occurs from high impact actions, high impact actions may still have low change in utility between states. A reframing that perhaps slightly simplifies the conservative AI problem is to constrain the agent to \textbf{not take high impact low (or highly uncertain) value actions}. For example, choosing to restrain a human that the model believes might commit a crime has debatable value, perhaps slightly positive, but is high impact and therefore should be avoided. The value-impact tradeoff is similar to the return-risk tradeoff, where we wish to maximize expected value within palatable amounts of risk. The whole motivation to care about low impact does stem from mitigating the risk of catastrophic actions afterall. It may be productive to explore this analogy further by designing objectives inspired by literature on the risk-reward tradeoff in economics. For example, its plausible that the agent shouldn't take actions that are not on the efficient frontier of the impact-value tradeoff curve.

\section{Further discussion and conclusion}

\subsection{Low-impact from MDP to POMDP}

Any approach to low-impact that ignores that impact depends on imperfect world models, that are subject to both refinement and tampering, may be limited in practice. Also, the approach should also consider that low-impact should depend on features of the world we care about, so the concept of low-impactedness should derive from values.

A possible framework that is still simplified, but do not ignore these problems is the POMDP/R setting $(\mathcal{S}, \mathcal{A}, T, \mathcal{O}, O)$, 
extending with an observation model $O : \mathcal{S} \to \Delta \mathcal{O}$. Now, we assume that the agent does \textit{not} have access to neither $T$ nor $\mathcal{O}$. Instead, letting $\mathcal{O}^{< \mathbb{N}}$ be the set of finite sequences of observations, the agent has access to an \textit{imperfect} probabilistic world model $\tilde{T}:\mathcal{O}^{<\mathbb{N}} \times \mathcal{A} \to \Delta \mathcal{O}$. The agent has then to use $\tilde{T}$ to both infer the baseline and the impact measure. The questions in this frame become:
\begin{itemize}
    \item Can the agent \textit{infer} an impact measure $d : \mathcal{O}^{<\mathbb{N}} \times \mathcal{O}^{<\mathbb{N}} \to \mathbb{R}$ from human data? Is this easier than inferring human reward? And what does a misspecified impact measure $\tilde{d}$ behave in presence of a true impact measure $d$?
    \item What agents can do to bypass this impact measure\footnote{An obvious answer is to tamper with its own impact measure model and world-model, so we already have to assume some restrictions here.}? And what a \textit{restricted} agent such as an oracle or an LLM can do to bypass these measures? And if not, can we prove it?
    \item How do impact measures behave under refinements $\tilde{T}^*$? In particular, how does splintering break our previous impact measure?
\end{itemize}

From this, some possible steps in a research agenda would be:
\begin{itemize}
    \item Reframing proposed approaches in the POMDP setting with an imperfect world model. Analyze, through a formal model of refinement and refactoring, how these approaches behave under refactoring and splintering.
    \item Analyzing whether those approaches are \textit{learnable} through human data. Don't ignore the fact that humans are biased and flawed, and have different preferences.
    \item Investigating, likely through some formal agency model, whether those approaches break with a capable enough agent. Moreover, study \textit{what} constraints should we assume to the AI, if those approaches don't break.
    \item Creating simplified toy world models that still obey the POMDP/R setting, and the imperfect world model assumption, and test approaches in these simplified toy world models. I think that we can simulate a lot in those world models.
\end{itemize}

\subsection{Low-impact agency and RLHF}

One may argue that progress in the has been constrained by debates over definitions which have each turned out to have some shortcomings.In fact, one issue with existing impact measures such as attainable utility is that they seem to conflate impact with utility. As discussed in Section~\ref{impactutility}, utility and impact are correlated but distinct concepts. Thus, we first need a way to measure impact independently from utilities. Reinforcement Learning using Human Feedback \cite{RLHF} has been proposed as a way to align models with human intuitions. Since we wish to define 'impact' consistent with human intutitions, RLHF may be a way to initially circumvent this problem altogether by making the model learn a fuzzy conception of 'impact' specific to the environment. Particularly, it seems useful to defer impact learning to finetuning once the model already has a world model which it can use to connect it's actions, their outcomes on the world, and how humans rated their impact. This circumvents the problem of updates to the world-model splintering the agent's learnt concept of impact as the finetuning can be designed to not change the world-model (one example is training a separate impact classifier, similar to the injury classifier \cite{RedWoodInjury}). 

ChatGPT can be seen as an example of using RLHF to constrain the agent from giving 'high impact' outputs. We can view the creators decision as constraining the goals of the model to being informative in a subset of conversations and defined the preferences learnt by the model, biases, ability to act as a romantic partner etc. as side effects. While correctly doing these may be valuable to some users, this is outweighed by the risk of negative impacts and thus ChatGPT simply states it is not designed to handle such conversations (noop action). This just shows that ChatGPT can be viewed as an instance of the value-impact framework, which is more general. Note that ChatGPT's success in preventing harmful outputs has been limited with numerous examples being constructed that circumvent safety checks. It is possible that a more general finetuning to avoid high-impact behaviours rather than the specific case-by-case (recommendations, bias, research production etc.) in ChatGPT can make it harder to circumvent safety checks. However, the model may simply learn to manipulate the low-impact check and exploit it's mis-specifications. Therefore, studies are needed to know to what extent this occur, and how this can be fixed. 

\subsection{Conclusion}
 Although low-impact agency seems like a promissing path toward safe AI, at least in the medium term, there are a lot theoretical and practical issues in not only implementing low-impactedness, but even defining it. This gives us the temptation to study simplified settings, where we can understand the problem more easily. However, we should not oversimplify the problem. We prefer to explicitly say that some solution depends on some part of the problem being solved, and pointing toward that part, than simply assuming the part away. After all, the final objective is to implement a robust low-impact approach in an AI system that will act in the real world, with all its complications, and not cause harm to it.

\clearpage
\bibliography{refs}
\bibliographystyle{plain}

\appendix

\end{document}